\pdfoutput=1

\documentclass[11pt]{article}

\usepackage[final]{acl}

\usepackage{times}
\usepackage{latexsym}
\usepackage{hyperref}

\usepackage[T1]{fontenc}

\usepackage[utf8]{inputenc}

\usepackage{microtype}

\usepackage{inconsolata}
\usepackage{enumitem}
\usepackage{graphicx}
\usepackage{booktabs}
\usepackage{multirow}
\usepackage{subcaption}
\usepackage{makecell}
\usepackage{caption}
\usepackage{float}
\usepackage{amsmath}
\usepackage{xcolor} 
\usepackage{tabularx}
\usepackage{tcolorbox}
\usepackage{minted}
\usepackage{algorithm}
\usepackage{algpseudocode}
\usepackage{xspace}
\usepackage{stfloats}
\usepackage{longtable}
\usepackage[nameinlink]{cleveref}
\usepackage[normalem]{ulem}
\useunder{\uline}{\ul}{}

\usepackage{listings}
\usepackage{xcolor}
\usepackage{graphicx}
\usepackage{subcaption}

\usepackage{tikz}
\usetikzlibrary{arrows.meta,positioning,fit}

\definecolor{codegreen}{rgb}{0.0,0.5,0.0}
\definecolor{codelightred}{rgb}{0.8,0.1,0.1}
\definecolor{borderblue}{RGB}{34,51,103}    
\definecolor{bggray}{RGB}{245,247,250}

\title{Interactive Training: \\Feedback-Driven Neural Network Optimization}

\author{Wentao Zhang \\
  University of Waterloo \\
  \texttt{w564zhan@uwaterloo.ca} \\\And
  Yang Young Lu \\
 University of Wisconsin-Madison\\
   \texttt{ylu97@wisc.edu} \\\And
  Yuntian Deng \\
   University of Waterloo \\
  \texttt{yuntian@uwaterloo.ca} \\}

\begin{document}
\maketitle
\begin{abstract}
Traditional neural network training typically follows fixed, predefined optimization recipes, lacking the flexibility to dynamically respond to instabilities or emerging training issues. In this paper, we introduce Interactive Training, an open-source framework that enables real-time, feedback-driven intervention during neural network training by human experts or automated AI agents. At its core, Interactive Training uses a control server to mediate communication between users or agents and the ongoing training process, allowing users to dynamically adjust optimizer hyperparameters, training data, and model checkpoints. Through three case studies, we demonstrate that Interactive Training achieves superior training stability, reduced sensitivity to initial hyperparameters, and improved adaptability to evolving user needs, paving the way toward a future training paradigm where AI agents autonomously monitor training logs, proactively resolve instabilities, and optimize training dynamics.

\end{abstract}


\section{Introduction}
Traditional neural network optimization typically involves setting hyperparameters and defining training strategies before execution, after which practitioners passively observe the training process until it completes or fails~\citep{bergstra2012random}. Despite its widespread adoption, this static training paradigm lacks flexibility and responsiveness once training begins. In practice, unforeseen challenges often arise mid-training, such as unstable loss dynamics, underperformance on specific tasks, or vanishing gradients in certain network components, all of which necessitate human intervention~\citep{takase2023spike,olmo20242}. Addressing these issues typically requires prematurely terminating the training job, manually adjusting hyperparameters or data configurations, and restarting the process~\citep{zhang2022opt}. On managed clusters, repeatedly resubmitting jobs exacerbates these inefficiencies, leading to wasted computational resources and delays due to job-queue overhead.



\begin{figure}[t]
\centering

\includegraphics[width=0.99\linewidth]{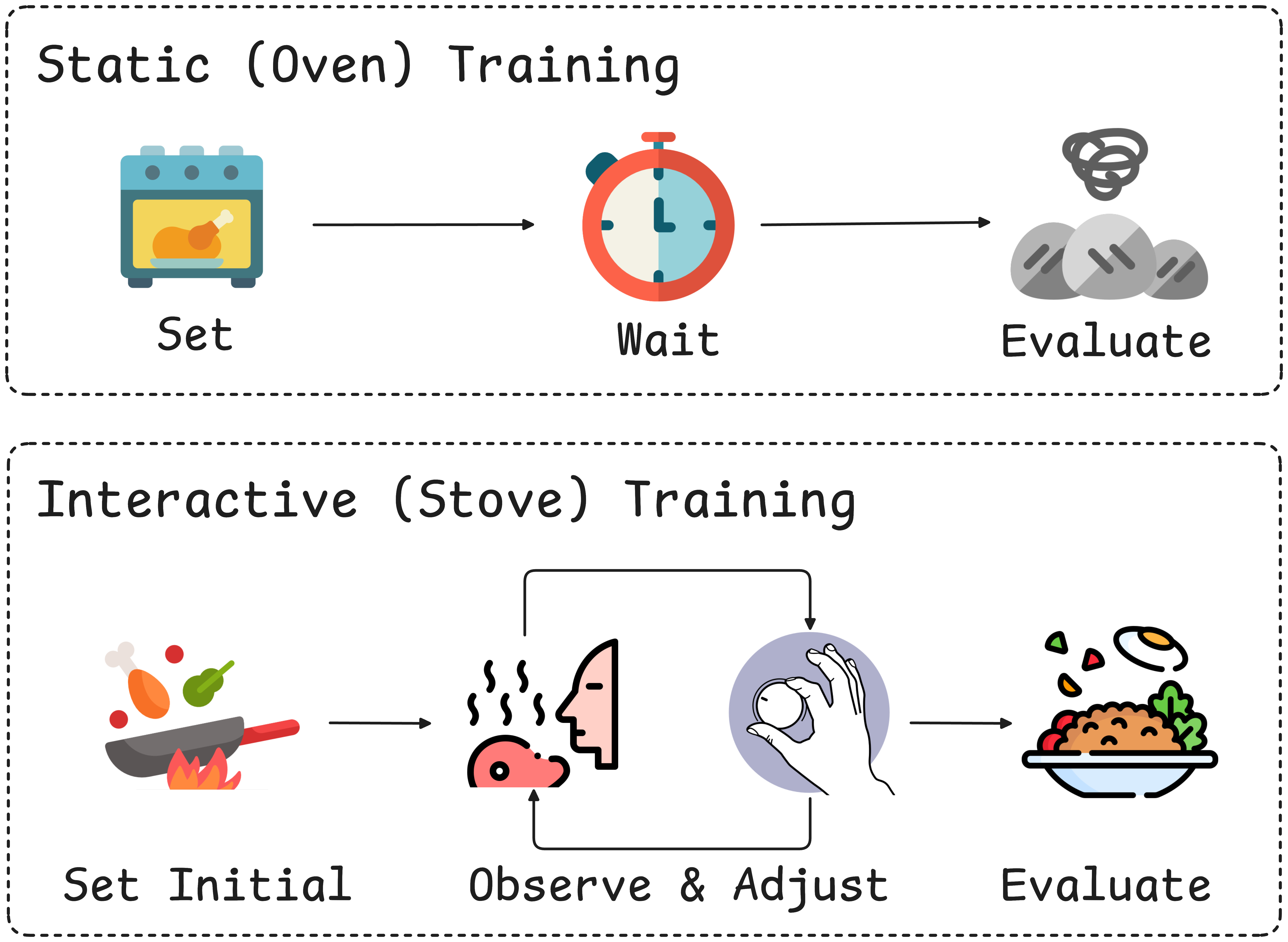}

\caption{Analogy of static vs. interactive training. Static training is like baking with a closed oven: hyperparameters follow a fixed, predetermined schedule until the end. Interactive training is like cooking on a stove: hyperparameters can be adjusted in real time.}
\label{fig:analogy-static-vs-interactive}
\end{figure}

In this paper, we introduce \emph{Interactive Training}, a framework enabling real-time feedback-driven optimization of neural networks, addressing the limitations of static training paradigms. Inspired by the intuitive act of adjusting a stove based on immediate sensory feedback during cooking (\Cref{fig:analogy-static-vs-interactive}), Interactive Training allows human experts or automated AI agents to dynamically intervene during training. Unlike traditional monitoring tools that only visualize training metrics, our approach transforms neural network optimization into an active and responsive process, enabling practitioners to continuously observe training progress, immediately react to emerging issues, and interactively guide the model toward improved outcomes.

Interactive Training enables users (human experts or automated AI agents) to dynamically adjust optimizer parameters, such as modifying learning rates in response to sudden spikes in loss. It supports mid-training updates to training data, allowing models to incorporate new data collected from real-world deployments without restarting training. Users can perform model-level interventions, such as reverting to previous checkpoints upon encountering unstable loss dynamics, or resetting specific parameters when invalid values are detected. The framework also provides gradient-level control, allowing users to dynamically set gradient clipping thresholds based on observed gradient norms, rather than relying on heuristic thresholds. 

\begin{figure*}[!t]
    \centering
    \includegraphics[width=0.99\linewidth, trim={0 0cm 18.5cm 0},clip]{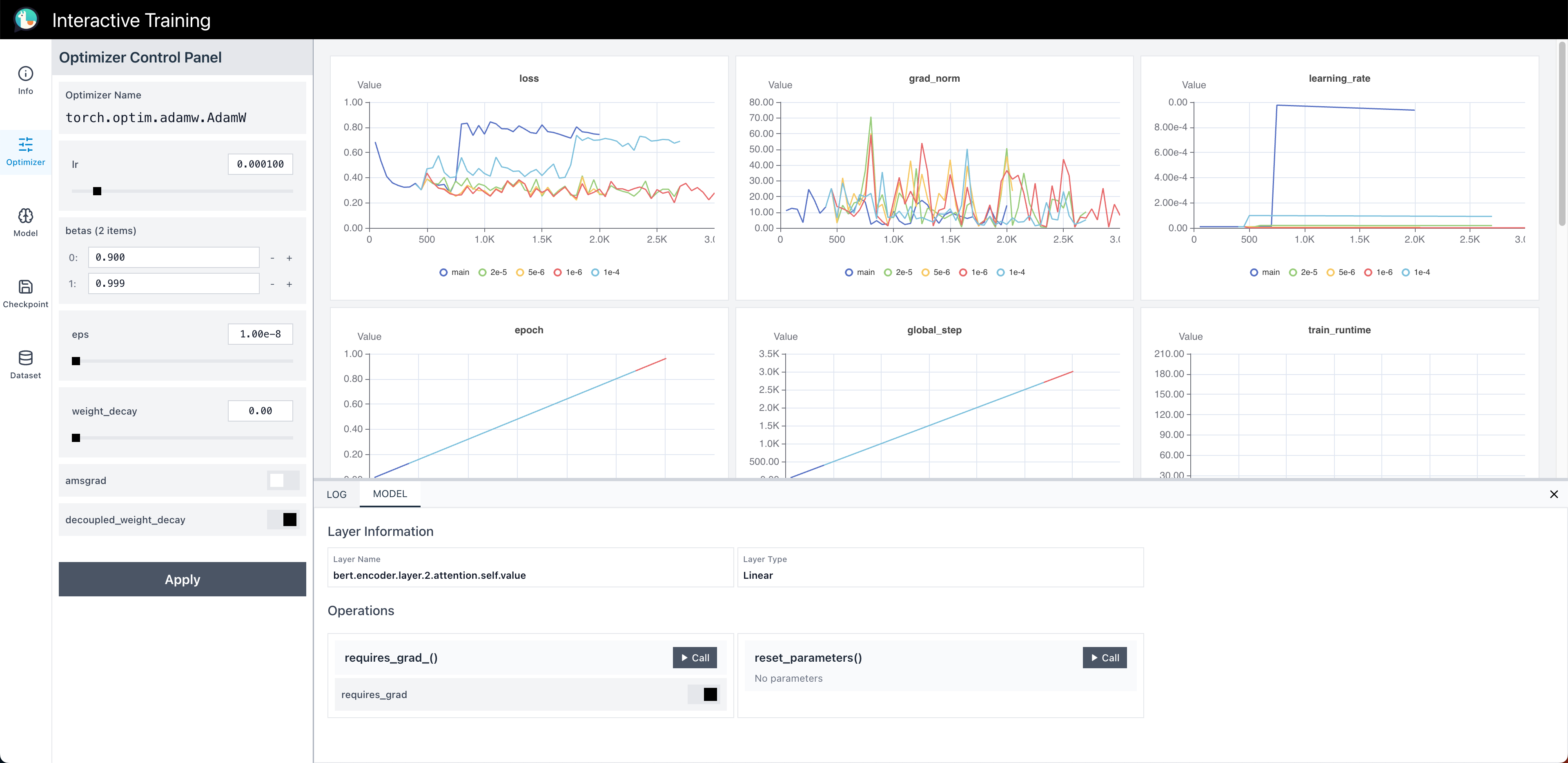}
    \caption{\label{fig:demo}Interactive Training Frontend Dashboard. The left panel provides control tabs organized by Optimizer, Model, Checkpoint, and Dataset, allowing users to dynamically send intervention commands during training (e.g., adjusting the learning rate via the Optimizer panel shown). The right side displays real-time visualizations of training metrics, such as loss and gradient norm. Unlike traditional monitoring tools, this interface supports active two-way communication, enabling users to directly intervene and influence ongoing training processes in real-time.}
\end{figure*}

We implement Interactive Training as an open-source library built on Hugging Face Transformers' widely adopted Trainer class~\citep{wolf-etal-2020-transformers}. At the core of our implementation is a control server acting as an intermediary between human experts (or AI agents) and the ongoing training process. This control server continuously listens on a predefined network port for incoming commands issued by users. Upon receiving commands (e.g., ``set learning rate to 1e-5''), it decodes these instructions into corresponding updates to optimizer parameters, model components, dataloaders, or gradients via callback functions invoked after each gradient step. The control protocol exposes its API endpoints through FastAPI. To facilitate ease of use for human experts, we also developed a React-based visualization dashboard, conceptually similar to Weights \& Biases~\citep{wandb}, which displays real-time training metrics across multiple plots (\Cref{fig:demo}). Crucially, unlike traditional monitoring tools, our frontend supports two-way communication: it not only visualizes training dynamics but also enables users to actively send control commands directly to the training loop.

We empirically validate Interactive Training through three case studies. First, we demonstrate that experienced human developers, leveraging real-time interactive adjustments, achieve superior optimization results compared to traditional static optimization methods on a language modeling task. Second, we showcase the potential for automated interventions by demonstrating that a general-purpose LLM-based AI agent, prompted with training logs, can autonomously correct suboptimal initial hyperparameters. Finally, we illustrate how our framework enables models to adapt in real-time to user-generated data collected during actual deployments~\citep{albalak2023efficient,wettig2025organize}, using a diffusion-based image generation application~\citep{ho2020denoising}. These studies collectively show exciting potential for human- and AI-driven Interactive Training.

Interactive Training transforms neural network optimization from a passive, static task into an active and responsive process. We envision a future in which model training is fully interactive, with hyperparameters, training data, and even loss functions dynamically adjusted based on mid-training feedback. Such interventions could be performed by human developers or, with even greater potential, by specialized automated AI agents designed explicitly to monitor training dynamics and evaluate intermediate model checkpoints. By bridging mid-training feedback with dynamic interventions, Interactive Training represents a paradigm shift toward continually improving neural network training workflows. To facilitate this vision, we have made our implementation openly available at \url{https://github.com/yuntian-group/interactive-training}, with an online demo accessible at \url{https://interactivetraining.ai}.

\begin{figure*}[!t]
    \centering
    \includegraphics[width=0.99\linewidth]{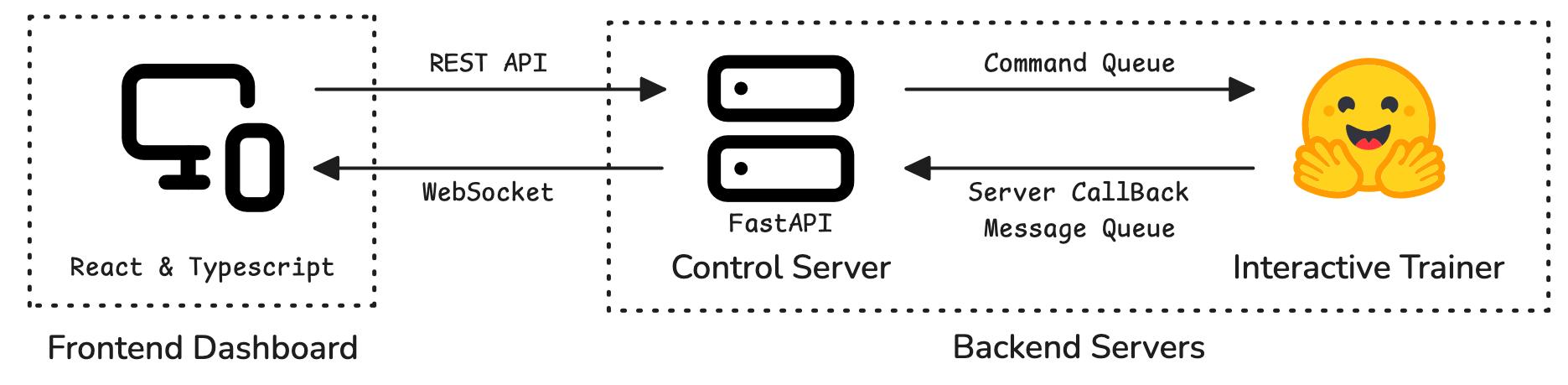}
    \caption{\label{fig:system}System Architecture. Users interact through a React-based Frontend Dashboard, which visualizes training metrics and sends control commands via REST API. The FastAPI-based Control Server mediates communication by forwarding user commands through command queues to the Interactive Trainer, implemented on top of Hugging Face's Trainer class. The trainer applies received commands via callback functions and sends real-time training updates back to the Control Server, which then broadcasts them to the Frontend Dashboard through WebSockets.}
\end{figure*}
\section{Interactive Training Framework}

\Cref{fig:system} provides an overview of our system architecture. At a high level, Interactive Training consists of three main components: a Control Server, which mediates communication between the trainer and users, managing commands, state updates, and training metrics; an Interactive Trainer, which performs model training and responds dynamically to intervention commands; and a Frontend Dashboard, which provides visualizations of training progress and enables users to issue real-time interventions.

\subsection{Control Server}
The Control Server acts as the central communication hub in Interactive Training, mediating interactions between the frontend dashboard and the interactive trainer. It serves two primary roles: receiving and dispatching user intervention commands, and broadcasting training updates back to clients.

Implemented using FastAPI, the Control Server exposes a set of APIs, allowing clients such as the frontend dashboard or automated AI agents to send intervention commands. Each command is represented as a JSON message specifying the action type (e.g., adjusting learning rates, checkpoint management) and its parameters (e.g., desired learning rate value). Upon receiving a command, the server enqueues it into command queues categorized by command type for asynchronous processing.

To enable real-time training updates, throughout training, the Interactive Trainer reports metrics such as loss values, gradient norms, and training status updates back to the server via event queues. The Control Server then broadcasts these updates to all subscribed clients, allowing users or automated agents to make timely intervention decisions.

Additionally, the server maintains state information such as training checkpoints, command history, and branched training logs. This state management not only supports reproducibility by logging each intervention but also enables interactive experimentation, such as reverting training to previous checkpoints or branching training trajectories.

The Control Server's modular design supports straightforward extensibility. We detail the currently supported intervention commands and discuss extensibility considerations in \Cref{sec:supported_apis}.

\subsection{Interactive Trainer}

The Interactive Trainer performs the actual model training, dynamically responding to intervention commands relayed from the Control Server. It extends Hugging Face's widely used \texttt{Trainer} class, augmenting it with callback functions that enable real-time interactivity without requiring significant changes to existing training scripts.

At its core, Interactive Trainer is implemented using custom callback functions passed to Trainer:

\begin{figure*}[!t]
  \centering
  \begin{lstlisting}[language=Python,
      basicstyle=\ttfamily\small,
      keywordstyle=\color{blue!60!black},
      commentstyle=\color{green!50!black},
      ]
from transformers import Trainer
from interactive_training import make_interactive # (1) Import helper

# (2) Wrap the standard Trainer class
InteractiveTrainer = make_interactive(Trainer)

# (3) Use them exactly as you would the original Trainer
trainer = InteractiveTrainer(...)

trainer.train()  # Training is now fully interactive!
  \end{lstlisting}
  \caption{\label{fig:interactive-code}Code changes required to enable Interactive Training.}
\end{figure*}

\begin{itemize}[leftmargin=*]
    \item \textbf{InteractiveCallback}: Handles runtime adjustments to training parameters, including optimizer hyperparameters, gradient clipping thresholds, and triggering on-demand model evaluations.
    \item \textbf{CheckpointCallback}: Saves/loads checkpoints.
    \item \textbf{LoggingCallback}: Captures training metrics, sending updates back to the Control Server after each gradient step.
    \item \textbf{RunPauseCallback}: Pauses/resumes training.
\end{itemize}

These callbacks communicate directly with the Control Server via dedicated command and event queues. Upon receiving commands from the Control Server (e.g., \texttt{``set learning rate to 1e-5''}), the respective callback updates the trainer's internal state at the next available gradient step, ensuring minimal disruption to the ongoing training loop.

In addition to callbacks for trainer control and metric logging, our framework also supports dynamic training data updates. We provide a function \texttt{make\_interactive\_dataset}, which can wrap PyTorch's \texttt{Dataset} and \texttt{IterableDataset} classes to make them controllable through user instructions.



Furthermore, the Interactive Trainer supports branching training trajectories. When reverting to earlier checkpoints, it can automatically create new branches of the training state, allowing multiple parallel or sequential training experiments to be compared. Each branch maintains its own isolated training history and checkpoints, providing a clear and reproducible record of experimentation paths.

\subsection{Frontend Dashboard}
The Frontend Dashboard (\Cref{fig:demo}) provides a user-friendly interface that enables users to visually monitor training progress and intervene in real-time. Built using React and TypeScript, the dashboard displays visualizations of key training metrics updated continuously via WebSocket connections established with the Control Server.

Unlike traditional monitoring dashboards that offer only passive visualizations, our frontend supports two-way communication. Users can dynamically issue intervention commands using intuitive control panels organized by intervention type (Optimizer, Model, Checkpoint, Dataset). Upon issuing a command, the frontend sends structured requests through RESTful API calls to the Control Server, which then communicates with the Interactive Trainer to apply the interventions at runtime.

Furthermore, the dashboard supports branched training trajectories, visualizing multiple experiment paths originating from a common checkpoint, allowing users to compare results from different settings. 

Additionally, the dashboard includes a log console at the bottom, which displays logs of each command issued, confirmation responses from the training process, as well as warnings and critical training events (e.g., ``Gradient overflow detected'').

\subsection{Usage Example}
To illustrate the simplicity of integrating Interactive Training into existing workflows, we highlight the minimal required modifications to a typical training script in \Cref{fig:interactive-code}. 
With minor adjustments, users immediately gain interactive control over training.

\section{Case Studies}
\begin{figure*}[t]
    \centering
    \begin{subfigure}[b]{0.48\textwidth}
        \centering
        \includegraphics[width=\textwidth]{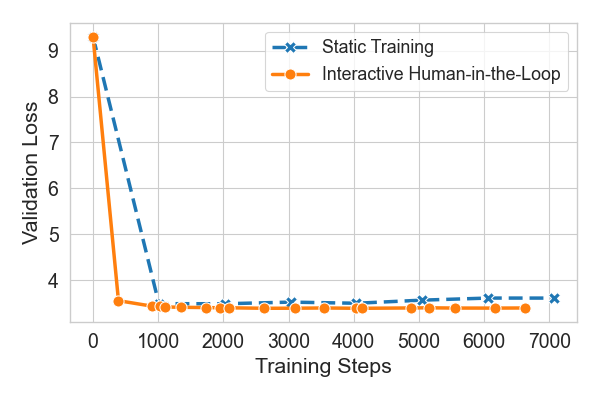}
        \caption{\label{fig:human-vs-static-loss}Validation loss curves}   
    \end{subfigure}
    \hfill
    \begin{subfigure}[b]{0.48\textwidth}
        \centering
        \includegraphics[width=\textwidth]{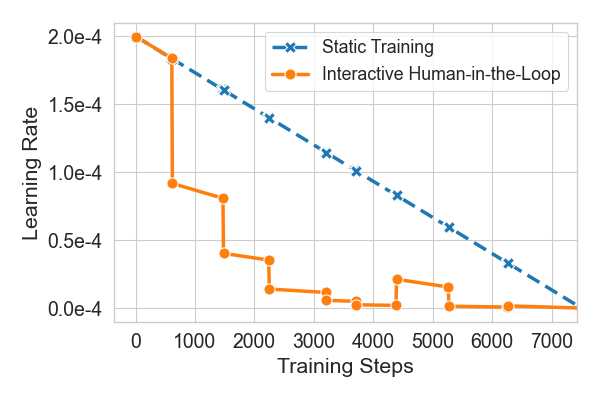}
        \caption{\label{fig:human-vs-static-lr}Learning rate schedules}
    \end{subfigure}
    \caption{
    \label{fig:human-vs-static}
        Comparison of human-in-the-loop Interactive Training versus traditional static training for finetuning GPT-2 on Wikitext-2. 
        \textbf{(a)}~Validation losses. Dynamic human interventions lead to improved optimization compared to the static baseline, which uses a fixed learning rate schedule.
        \textbf{(b)}~Actual learning rates used over steps.
    }
\end{figure*}
\subsection{Human-in-the-Loop Intervention}
We first demonstrate the benefits of human-in-the-loop Interactive Training by finetuning GPT-2~\citep{radford2019language} on Wikitext-2~\citep{merity2017pointer}. Our goal is to evaluate whether human interventions could yield improved optimization results. 

\paragraph{Baseline}
For the baseline, we trained the model using a fixed learning rate schedule, starting with an initial learning rate of $1\times10^{-5}$ and linearly annealing it to zero over the entire training duration.

\paragraph{Human Intervention}
The Interactive Training setup mirrored the baseline, except that a human expert dynamically adjusted the learning rate based on training dynamics visualized in the dashboard.
 
\paragraph{Results}
 \Cref{fig:human-vs-static-loss} compares the results of both approaches. The interactive method achieves lower validation losses than the static baseline. By inspecting the learning rate schedules (\Cref{fig:human-vs-static-lr}), we find that the human expert effectively responded to the model's real-time behavior. For instance, upon observing training loss oscillation resulting from an initially high learning rate, the expert reduced the learning rate, improving convergence.

\subsection{LLM-in-the-Loop Intervention}
Next, we investigate the feasibility of automating training interventions by leveraging an AI agent. Specifically, we evaluate whether an LLM, provided with training logs, can correct training instabilities caused by suboptimal hyperparameters.

\paragraph{Setup}
We follow the same setup as in the previous study but deliberately introduce instability by initializing the training with an excessively large learning rate ($5\times10^{-3}$) and disabling the learning rate scheduler. This excessively high learning rate causes poor convergence of the training.

Instead of human interventions, we introduce an automated LLM-based agent, using OpenAI's o4-mini model~\citep{openai_o4mini_2025}. At every step, the LLM agent receives a textual summary of recent training logs---including current and historical training losses, validation losses, learning rates, and step counts---and is prompted to determine the next action regarding the learning rate (doubling, halving, or keeping it unchanged). The detailed prompt template is provided in \Cref{sec:appendix_llm_prompt}. 



\begin{figure*}[!t]
    \centering
    \begin{subfigure}[b]{0.48\textwidth}
        \centering
        \includegraphics[width=\textwidth]{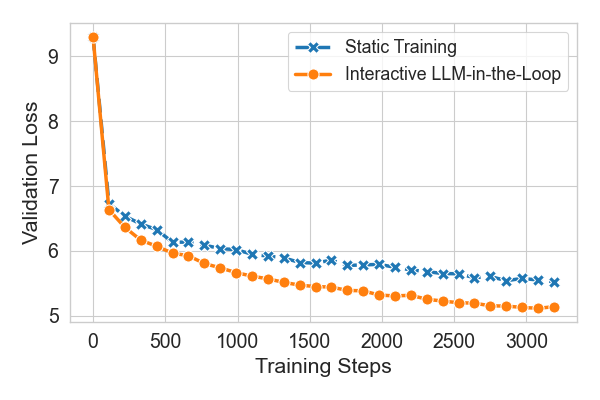}
        \caption{\label{fig:llm-vs-static-loss}Validation loss curves}   
    \end{subfigure}
    \hfill
    \begin{subfigure}[b]{0.48\textwidth}
        \centering
        \includegraphics[width=\textwidth]{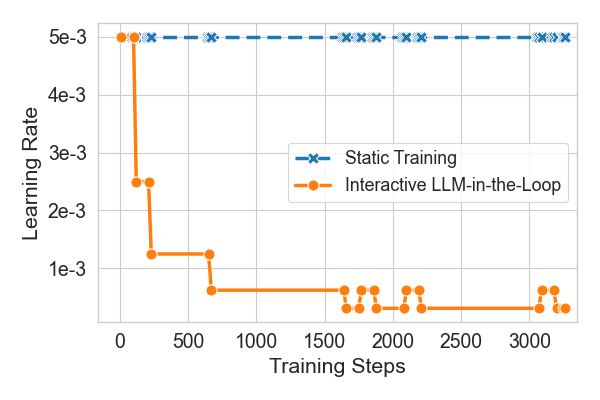}
        \caption{\label{fig:llm-vs-static-lr}Learning rate schedules}
    \end{subfigure}
    \caption{\label{fig:llm-vs-static}
        Comparison of LLM-in-the-loop automated intervention versus static training with a fixed, excessively large learning rate. 
        \textbf{(a)} Validation losses. LLM-based intervention effectively stabilizes optimization. 
        \textbf{(b)} Learning rate trajectory. Initially high learning rate is reduced by the LLM agent in response to observed loss instabilities.
    }
\end{figure*}

\paragraph{Results}
The results of this automated intervention approach are in \Cref{fig:llm-vs-static}. LLM-in-the-loop training recovers from the initial suboptimal learning rate by recommending timely reductions. This study demonstrates the potential of using AI agents to automate Interactive Training interventions.



\subsection{Real-time Training Data Updates}
Finally, we demonstrate how Interactive Training enables continuous model improvement through dynamic updates to training data collected from real-world deployments. We apply our framework to NeuralOS~\citep{rivard2025neuralos}, which uses a diffusion model to simulate a real operating system by predicting the next screen frame given user mouse and keyboard inputs. After deploying the initial NeuralOS model online at \url{https://neural-os.com}, we continuously collected real user interactions. 

\paragraph{Setup}
Initially, the NeuralOS model was trained for two months on a large synthetic dataset generated from scripted interactions. After deployment, we gathered 746 demonstration sequences (88K frame transitions) from real user interactions over a period of 14 days. Using the Interactive Training framework, we dynamically updated the training data of a continuously running finetuning process, incorporating newly collected data on-the-fly. Additionally, model checkpoints during finetuning were continuously uploaded, updating the deployed model to reflect improvements in real-time.

\paragraph{Results}
After dynamically finetuning NeuralOS, we observed substantial improvements, especially for tasks frequently performed by actual users, such as interacting with the Firefox browser and creating new folders. Representative examples illustrating the significant improvement achieved by incorporating real user data are shown in \Cref{fig:firefox_and_newfolder} (\Cref{sec:appendix_case_neuralos}). This case study demonstrates that Interactive Training effectively enables deployed models to adapt to real-world usage patterns.

\section{Limitations}
\paragraph{Reproducibility}
Different experts or AI agents training the same model might perform different interventions, leading to different outcomes. While we acknowledge this inherent variability, it is worth noting that large-scale neural network training today already involves substantial expert intervention. For instance, Meta's OPT language model required at least 35 manual restarts due to hardware failures and involved manually selecting checkpoints to recover from loss divergences~\citep{zhang2022opt}. To mitigate reproducibility concerns, our implementation logs all interventions, enabling replay.


\paragraph{Expertise Requirement}
Interactive Training requires human experts or automated agents to possess expertise to identify appropriate intervention points and apply effective corrective actions. AI agents may lack adequate demonstrations in their training data to reliably intervene, due to the novelty of this optimization paradigm. However, we view this limitation as an opportunity, motivating future research into specialized intervention agents.

\section{Future Work}

\paragraph{Feedback-Driven Data Adjustment}
Real-time interventions could enable new optimization strategies. Users or AI agents could periodically evaluate intermediate checkpoints to identify model weaknesses and then dynamically adjust training data accordingly, either by injecting targeted synthetic examples, or by adjusting data mixture weights to emphasize relevant existing examples. 
 
\paragraph{Training Health Diagnostic Metrics}
Just as periodic checkups help humans address potential health issues, model training could benefit from analogous health monitoring metrics. One promising direction is to develop ``health indicators'' such as the standard deviation of hidden states across training examples to detect ``dead'' neurons~\citep{ioffe2015batch}. More sophisticated analyses~\citep{hu2023latent} could also provide signals prompting human or AI interventions.


\paragraph{AI Agents for Training Intervention}
Finally, we envision a future where AI agents autonomously monitor health indicators and proactively intervene to improve training stability and efficiency. While this paper demonstrates simple log-based prompts to a general-purpose LLM, specialized intervention agents explicitly trained to detect anomalies and guide training represent a promising direction. 

\section{Related Work}
\paragraph{Human-in-the-Loop Machine Learning} A rich body of work has explored human interventions during model training. In active learning, the learning algorithm remains in control but queries human annotators for labels on selected examples~\citep{mosqueira2023human}. Interactive machine learning goes further by allowing human feedback beyond just labeling, such as by correcting predictions or adjusting inputs~\citep{fails2003interactive}. Another paradigm, machine teaching, gives human domain experts explicit control over the training process, such as by designing the sequence or structure of tasks to transfer knowledge to the model~\citep{simard2017machine}. These approaches demonstrate the value of human insight during training; however, they often rely on predefined schedules or specific forms of input rather than truly real-time, open-ended intervention. Our Interactive Training framework aims to allow humans (or AI agents) to intervene training at any moment, which extends human-in-the-loop learning from static plans to live control.

\paragraph{Automated ML and Adaptive Optimization} Orthogonal to human guidance, AutoML research has developed methods to automate hyperparameter tuning and training optimization. Traditional approaches include Bayesian optimization and bandit strategies that adaptively select hyperparameter configurations across trial runs~\citep{li2018hyperband}. More recent techniques seek to adapt within a single run: learning rate scheduling is routinely used to vary the step size during training, and researchers have even applied reinforcement learning to discover optimized scheduling policies automatically~\citep{Subramanian2023,xu2019learning}. Also related to our work, Population-Based Training (PBT)~\citep{jaderberg2017population} learns an automatic dynamic schedule of hyperparameters. Interactive Training complements AutoML by enabling both automated agents and human experts to adjust training trajectories in real time. Instead of treating training as a black-box process to tune from the outside, our framework opens the loop, so scheduling decisions and hyperparameter tweaks can occur on the fly, guided by live signals or human judgment.

\paragraph{AI Agents for Training Control and Debugging} Researchers have started to consider AI agents as participants in the training loop. For example, \citet{epperson2025interactive} developed an interactive debugger for multi-agent AI workflows that allows a user to reset agents to earlier states and alter their messages mid-execution. Modern visualization platforms are beginning to integrate automated agents to monitor experiment runs, detect anomalies, and even suggest hyperparameter adjustments based on the accumulated training data~\citep{relevanceai_weights-biases}. However, such agents typically remain advisory tools; they do not directly plug into the training loop to enact immediate interventions. Our Interactive Training framework builds on this idea by permitting both humans and AI agents to not only analyze but also modify a running training job. This bridges a gap between AI-driven monitoring and actual training control, turning insights into on-the-fly actions.

\section{Conclusion}
We presented Interactive Training, a framework that reimagines neural network training as an interactive, feedback-driven process, with either humans or AI agents dynamically controlling training strategies mid-training. Through real-time interventions, Interactive Training enables adjusting optimization parameters, training data, and model components on-the-fly based on insights gained during training. Case studies show advantages over traditional static training paradigms: improved accuracy, reduced sensitivity to initial hyperparameters, and real-time adaptation to evolving application needs.

Interactive Training introduces a new dimension to training workflows: responsiveness. Just as modern software development evolved from rigid release cycles toward agile and continuous integration practices, we advocate for a parallel shift in neural network optimization. The training process need not remain a static black box where practitioners must passively await results. Instead, it can become an interactive process, allowing continuous monitoring and intervention based on emerging information and feedback. We have open-sourced our framework, inviting the community to provide feedback and contribute to further development. 


\section*{Acknowledgements}
Yuntian Deng acknowledges support from an NSERC Discovery Grant (RGPIN-2024-05178), a Starter Grant from the University of Waterloo, and research funding from Manulife. Wentao Zhang is supported in part by these sources and by the Dr. Derick Wood Graduate Scholarship, generously funded by Ms. Mary Chen.

\bibliography{custom}
\appendix

\clearpage

\section{\label{sec:supported_apis}Supported Interactive Commands}
Interactive Training supports real-time intervention through structured commands. We first describe the general command message format, followed by details of supported commands grouped by category.

\subsection{Command Message Format}

All commands follow a uniform JSON message structure:

\begin{verbatim}
{
  "command": "[command_name]",
  "args": "[command_arguments_as_json]",
  "time": [unix_timestamp],
  "uuid": "[unique_identifier]",
  "status": "[status]"
}
\end{verbatim}

\begin{itemize}[leftmargin=*]
\item \texttt{command}: Type of command (as listed below).
\item \texttt{args}: JSON-formatted arguments specific to the command.
\item \texttt{time}: UNIX timestamp indicating when the command was issued.
\item \texttt{uuid}: Unique identifier for tracking command status.
 \item \texttt{status}: Current execution state, which can be one of: \texttt{"requested"}, \texttt{"pending"}, \texttt{"running"}, \texttt{"completed"}, \texttt{"success"}, or \texttt{"failed"}.\end{itemize}

\subsection{Supported Commands}

Interactive Training currently supports the following real-time intervention commands, organized by their intended use:

\paragraph{Optimizer Adjustment}
\begin{itemize}[leftmargin=*]
\item \texttt{update\_optimizer}: Adjust optimizer hyperparameters (e.g., learning rates, momentum, weight decay) during training.
\end{itemize}

\textbf{Example}:
\begin{verbatim}
{
  "command": "update_optimizer",
  "args": "{\"lr\": {\"value\": 1e-5}}"
}
\end{verbatim}

\paragraph{Checkpoint Management}
\begin{itemize}[leftmargin=*]
\item \texttt{save\_checkpoint}: Save the current model state as a checkpoint.
\item \texttt{load\_checkpoint}: Load a previously saved checkpoint and optionally branch training from that point.
\end{itemize}

\textbf{Example}:
\begin{verbatim}
{
  "command": "load_checkpoint",
  "args": "{\"uuid\": \"[uuid]\"}"
}
\end{verbatim}

\paragraph{Training Control}
\begin{itemize}[leftmargin=*]
\item \texttt{pause\_training}: Pause the training loop.
\item \texttt{resume\_training}: Resume training after being paused.
\item \texttt{stop\_training}: Terminate the training process immediately.
\end{itemize}

\paragraph{Model Interventions}
\begin{itemize}[leftmargin=*]
\item \texttt{model\_layer\_operation}: Run a method of a layer such as resetting or reinitializing specified model parameters (e.g., upon detecting NaN values or activation collapse).

\item \texttt{model\_layer\_parameter\_update}: Update layer hyper-parameter, e.g. dropout value of a dropout layer. 

\end{itemize}

\paragraph{Dataset Management}
\begin{itemize}[leftmargin=*]
\item \texttt{update\_dataset}: Update training data mid-training, e.g., to incorporate newly collected user data.

\item \texttt{update\_dataset\_runtime\_hyperparameters}: Update dataset run time hyper-parameters, e.g. the mixing ratio of different part or subset of datasets. 
\end{itemize}

\paragraph{Evaluation and Monitoring}
\begin{itemize}[leftmargin=*]
\item \texttt{do\_evaluate}: Trigger a model evaluation step on the validation dataset.
\end{itemize}

\subsection{Extensibility}
Our framework is designed for easy extensibility. New commands and interactions can be added by defining new command types, registering handlers, and extending the control server and trainer callbacks. We encourage community contributions of new intervention commands and interaction patterns via our open-source repository.

\section{\label{sec:appendix_llm_prompt}Detailed LLM Prompt}

The exact textual prompt provided to the LLM-based intervention agent for automated learning rate adjustments is shown in \Cref{fig:prompt_llm_lr_adjustment}. At each intervention point, the placeholders such as \texttt{{{current\_step}}}, \texttt{{{current\_lr}}}, \texttt{{{lr\_history}}}, \texttt{{{train\_loss\_history}}}, and \texttt{{{valid\_loss\_history}}} are dynamically replaced with the most recent training data and metrics before the prompt is sent to the LLM agent. The agent is explicitly instructed to respond with a clear JSON-formatted decision---choosing either to double, halve, or keep the learning rate unchanged---accompanied by a brief explanation within 100 words.

\begin{figure*}[htbp]
    \centering
    \begin{tcolorbox}[colframe=borderblue, colback=bggray, left=1mm, right=1mm, top=0.5mm, bottom=0.5mm]
\begin{minted}{markdown}
You are an expert in language model tuning. You are going to adjust the learning rate of fine-tuning a GPT-2 model using the WikiText-2 train data with a constant learning rate scheduler. Your goal is to minimize the final validation loss.

# Log History

current step: {{current_step}}

current learning rate: 

{{current_lr}}

learning rate history: 

{{lr_history}}

train loss history: 

{{train_loss_history}}

validation loss history: 

{{valid_loss_history}}


# Instruction

- Based on the log history, decide how to change the learning rate. The log history includes metrics like validation loss, training loss, and epoch count.

- You MUST output choose between following three actions: 
    1. "Double", double the learning rate 
    2. "Half", reduce the learning rate to 50% of the learning rate. 
    3. "Same", make no change.

- Respond in JSON format with an explanation within 100 words:

{
    "explanation": "<Explanation in 100 words>",
    "action": "<'Double', 'Half', or 'Same'>"
}

# Response
\end{minted}
    \end{tcolorbox}
    \caption{Prompt used for the LLM-based automated learning rate adjustment.}
    \label{fig:prompt_llm_lr_adjustment}
\end{figure*}

\begin{figure*}[t!]
    \centering
    \setlength{\tabcolsep}{2pt}
    \renewcommand{\arraystretch}{0.5}
    \begin{tabular}{cccc}
    Pred. Frame (early) $\cdots$ & Pred. Frame $\cdots$ & Pred. Frame $\cdots$ & Pred. Frame (late)\\[10pt]

    \multicolumn{4}{c}{\textbf{Firefox Interaction (Before Real-Time Training Data Updates, Unsuccessful)}}\\[4pt]
        \includegraphics[width=0.24\linewidth]{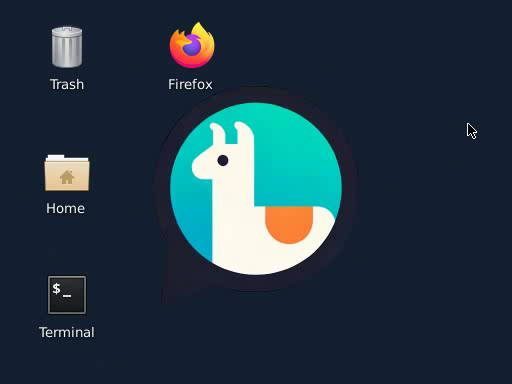} &
        \includegraphics[width=0.24\linewidth]{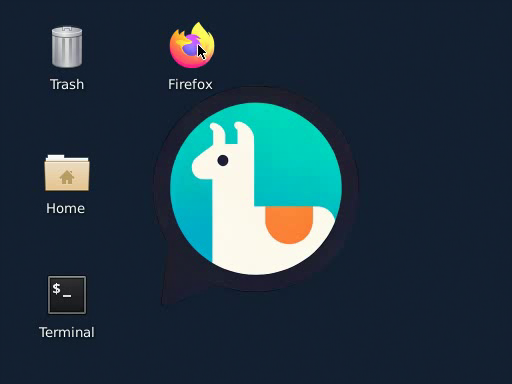} &
        \includegraphics[width=0.24\linewidth]{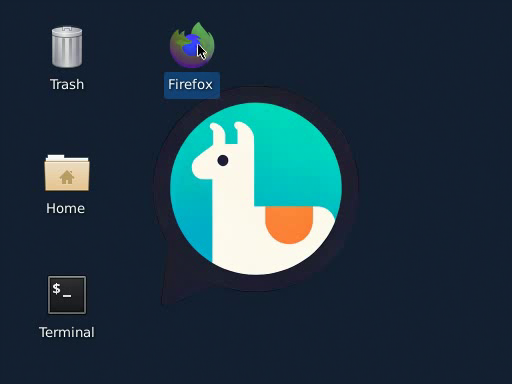} &
        \includegraphics[width=0.24\linewidth]{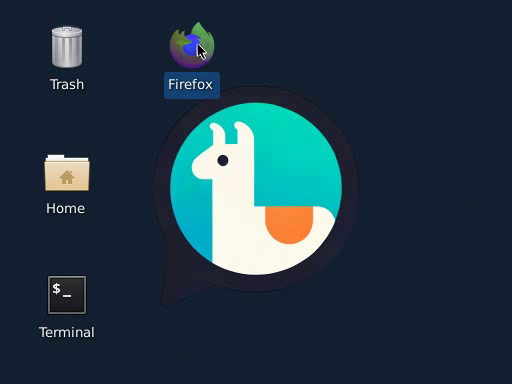} \\[8pt]

    \multicolumn{4}{c}{\textbf{Firefox Interaction (After Real-Time Training Data Updates, Successful)}}\\[4pt]
        \includegraphics[width=0.24\linewidth]{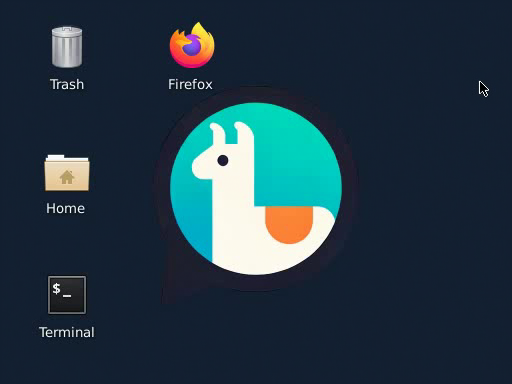} &
        \includegraphics[width=0.24\linewidth]{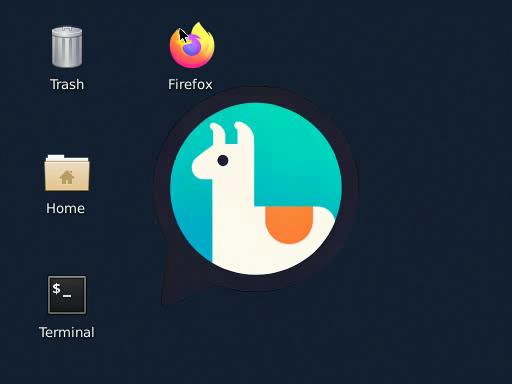} &
        \includegraphics[width=0.24\linewidth]{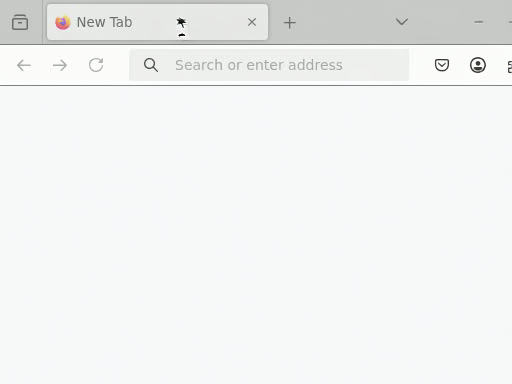} &
        \includegraphics[width=0.24\linewidth]{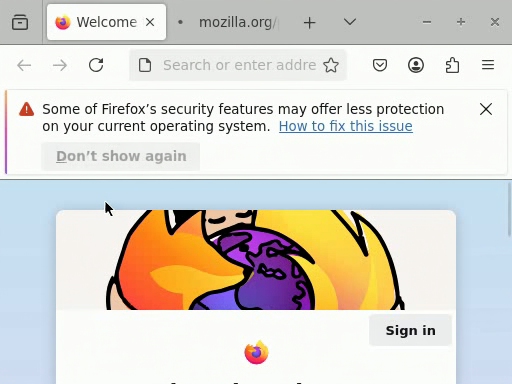} \\[40pt]

    \multicolumn{4}{c}{\textbf{Folder Creation (Before Real-Time Training Data Updates, Unsuccessful)}}\\[4pt]
        \includegraphics[width=0.24\linewidth]{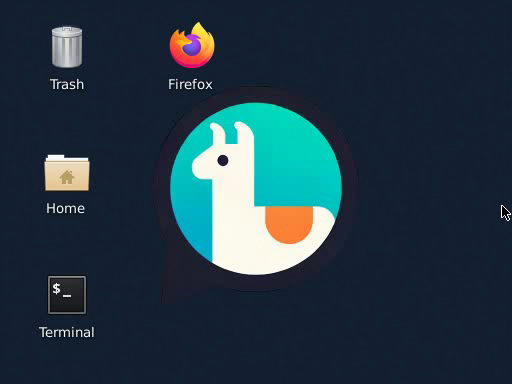} &
        \includegraphics[width=0.24\linewidth]{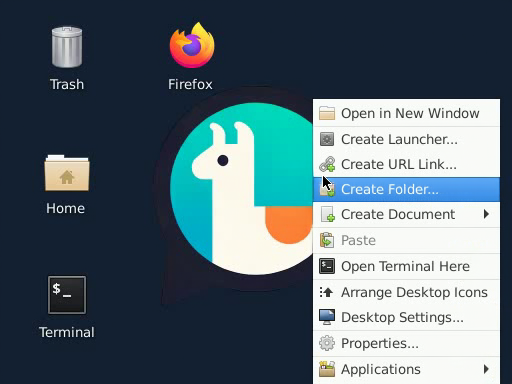} &
        \includegraphics[width=0.24\linewidth]{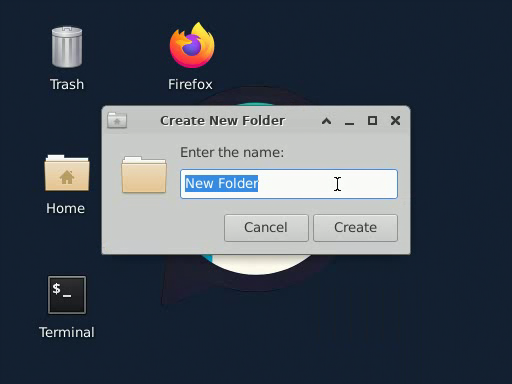} &
        \includegraphics[width=0.24\linewidth]{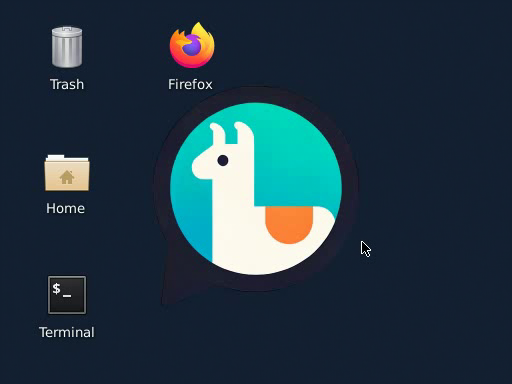} \\[8pt]

    \multicolumn{4}{c}{\textbf{Folder Creation (After Real-Time Training Data Updates, Successful)}}\\[4pt]
        \includegraphics[width=0.24\linewidth]{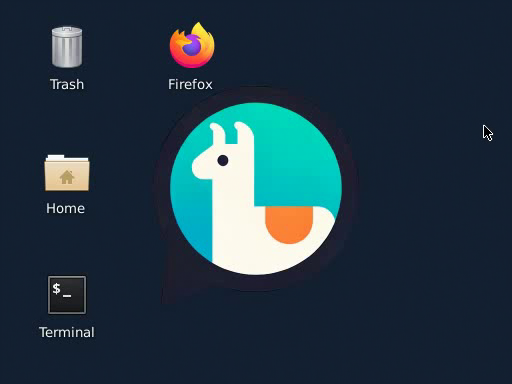} &
        \includegraphics[width=0.24\linewidth]{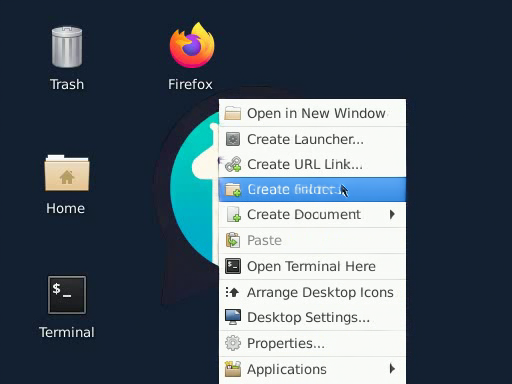} &
        \includegraphics[width=0.24\linewidth]{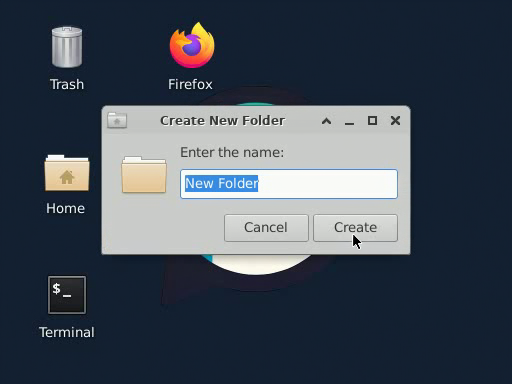} &
        \includegraphics[width=0.24\linewidth]{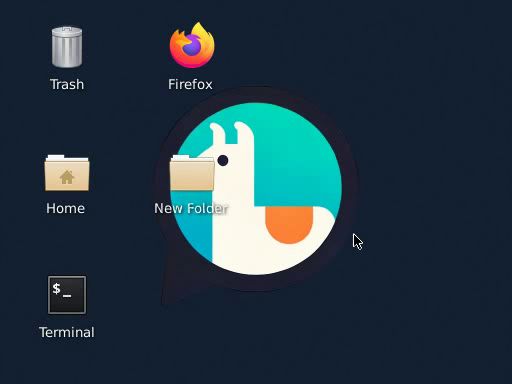} \\
    \end{tabular}
    \caption{Qualitative comparison illustrating improvements in the NeuralOS model's behavior before and after real-time training data updates using data collected from real users. The first two rows demonstrate model predictions when launching the Firefox browser, while the last two rows demonstrate creating a new folder. For each task, the top row is before finetuning and the bottom row after finetuning.}
    \label{fig:firefox_and_newfolder}
\end{figure*}

\section{\label{sec:appendix_case_neuralos}Results for NeuralOS Case Study}
\Cref{fig:firefox_and_newfolder} shows representative qualitative comparisons illustrating the improvements obtained by finetuning the NeuralOS model with real-time training data updates from actual user interactions collected through an online demo at \url{https://neural-os.com}. The figure consists of four rows: the top two rows depict model behavior when interacting with the Firefox browser, comparing performance before (first row) and after (second row) interactive finetuning. Before finetuning, attempts to open Firefox usually failed, leaving the screen on the desktop. This is because opening Firefox is challenging to predict, as the browser takes a longer time to launch compared to other applications (often more than 40 frames after clicking). After finetuning, however, opening Firefox is typically successful, due to frequent occurrences of Firefox interactions in the collected real user data. Similarly, the bottom two rows demonstrate the model's improved capability in creating new folders, comparing behavior before (third row, unsuccessful) and after (fourth row, successful) incorporating real user data. These examples highlight how Interactive Training effectively enables the model to naturally adapt to tasks frequently attempted by real users.

\end{document}